\documentclass[]{onedayagent}
\usepackage{fix-cm}
\usepackage{helvet}

\usepackage{xcolor}
\usepackage{multirow}
\usepackage{colortbl}
\usepackage{tabularx}
\usepackage{tcolorbox}
\usepackage{enumitem}
\usepackage{pifont}
\usepackage{wrapfig}
\usepackage{url}

\definecolor{mygray}{gray}{0.9}
\definecolor{myblue}{HTML}{F0FFFF}
\definecolor{myblue_light}{HTML}{EAF6FF}
\definecolor{myblue_dark}{HTML}{2B57A0}
\definecolor{myblue_lightv1}{HTML}{F2F7E0}
\definecolor{mygreen}{RGB}{60, 179, 113}
\definecolor{myyellow_light}{RGB}{255, 255, 224}
\definecolor{lightgreen}{HTML}{D3E5C7}
\definecolor{deepgreen}{RGB}{46,139,87}

\newcommand\blfootnote[1]{%
  \begingroup
  \renewcommand\thefootnote{}\footnote{#1}%
  \addtocounter{footnote}{-1}%
  \endgroup
}

\title{OneDayAgent: Towards a Long-Horizon Harness for Autonomous Agents}

\author[1,2]{Jingsheng Zheng}
\author[3]{Xinyuan Fang}
\author[1,2]{Jintian Zhang}
\author[2]{Zhengke Gui}
\author[1]{Huajun Chen}
\author[1\,\dag]{Ningyu Zhang}

\affiliation[1]{\mbox{Zhejiang University}}
\affiliation[2]{\mbox{Ant Group}}
\affiliation[3]{\mbox{Independent Researcher}}

\abstract{
LLM agents are increasingly applied to open-ended everyday requests that span work, study, and life.
These tasks are long-horizon, cross-environment, and multimodal, forcing the agent to preserve goals and constraints across many steps while navigating heterogeneous tools and attachments.
While prior work has addressed individual failure modes such as goals drift, states loss, and context overflow, whether a single harness can manage them jointly and remain effective across backends has received less study. We present OneDayAgent, a long-horizon harness for autonomous agents. OneDayAgent turns an open-ended request into a managed execution process that decomposes tasks into bounded subtasks, maintains execution memory under context pressure, and verifies and repairs the final deliverable. We evaluate OneDayAgent on AgentIF-OneDay across 104 tasks. With the GLM-5.2 backend, OneDayAgent sets a new state of the art with an overall score of 0.821. The same harness runs across five backend LLMs from three model families, indicating the harness generalizes across backends without tuning, even as different models induce distinct execution styles under the same workflow.
}

\badge{\faHome}{Homepage}{https://github.com/zjunlp}
\badge{\faGithub}{Code}{https://github.com/zjunlp}
\badge{\faDatabase}{Data}{https://github.com/zjunlp}
\badge{\faHuggingFace}{Model}{https://huggingface.co/zjunlp}
\badge{\faEnvelope}{Contact}{mailto:zhangningyu@zju.edu.cn}

\begin{document}

\blfootnote{$^\dagger$Corresponding author.}
\maketitle

\begin{figure}[t]
    \centering
    \includegraphics[width=1.0\textwidth]{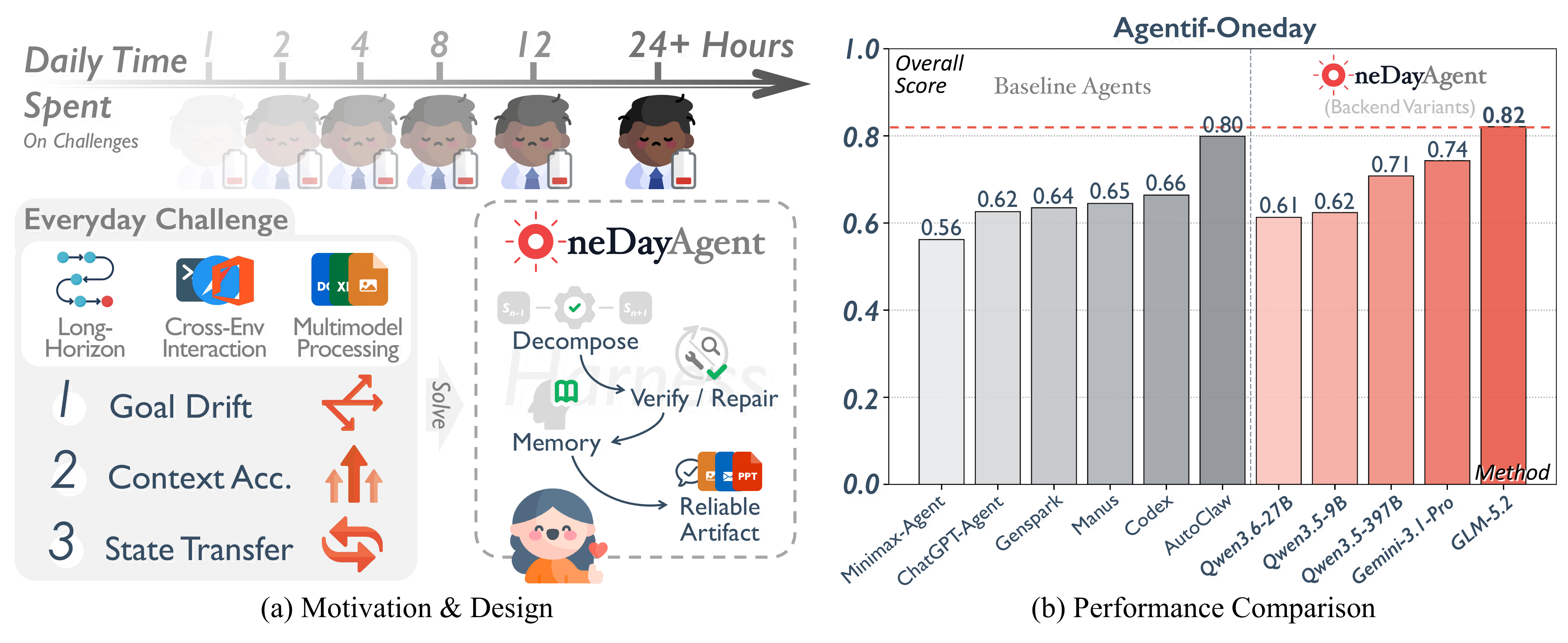}
    \caption{\textbf{Long-Horizon Everyday Tasks.} \textbf{(a)} Everyday requests have long-horizon, cross-environment, and multimodal characteristics. These create goal drift, context accumulation, and state transfer failures. OneDayAgent addresses them through task decomposition, verification and repair, and execution memory to produce reliable deliverables. \textbf{(b)} OneDayAgent achieves a new state of the art on AgentIF-OneDay with an overall score of 0.821.}
    \label{fig:intro}
\end{figure}

\section{Introduction}

Large language models are increasingly deployed as agents for tasks beyond single-turn question answering~\cite{ferrag2026llmreasoningautonomousai}, ranging from software engineering~\cite{tang2026llmbasedagenticsystemssoftware} and computer use~\cite{Sager_2026} to deep research~\cite{yu2026deepresearchdeepresearch} and personalized assistance~\cite{xu2026personalizedllmpoweredagentsfoundations}.
A growing share of real-world agent usage, however, involves open-ended everyday requests that span work, study, and life. A single instruction may require collecting web evidence, editing local files, and producing a deliverable such as a deck or report.
Unlike short tasks, these requests are \textbf{long-horizon}, \textbf{cross-environment}, and \textbf{multimodal} (Figure~\ref{fig:intro}(a)), forcing the agent to preserve goals and constraints across many steps while navigating heterogeneous tools and attachments.
Benchmarks such as AgentIF-OneDay~\cite{chen2026agentifonedaytasklevelinstructionfollowingbenchmark} formalize this shift by evaluating task-level instruction following with concrete deliverables.

As the horizon grows from minutes to hours, multi-step decision-making, unlike retrieval or temporal prediction, faces sustained pressure from context accumulation. \textbf{Goals drift} from accumulated constraints and \textbf{intermediate state fails to transfer} across environments. For example, an agent that first researches a topic on the web and later edits a local deliverable may drop an early formatting requirement by the time it reaches the editing step, or lose the search evidence gathered in an earlier subtask when it switches to the file environment, so the final artifact omits content that was already found. Existing approaches address individual failure modes through reasoning scaffolds, feedback-based revision, or memory management, but these failures interact and compound, so fixing one in isolation does not suffice.

We present OneDayAgent, a long-horizon harness that turns an open-ended request into a managed execution process built on three capabilities. Task decomposition breaks an overloaded request into bounded subtasks, global verification and repair re-aligns the deliverable with the original intent and patches localized defects, and execution memory compresses observations and checkpoints subtask state under context pressure. All capabilities operate over a unified action space covering web, computation, file, and multimodal tools. On AgentIF-OneDay across 104 tasks, the GLM-5.2 backend~\cite{glm5team2026glm5vibecodingagentic} achieves a \textbf{new state of the art} with an overall score of \textbf{0.821} (Figure~\ref{fig:intro}(b)), leading across all task types, domains, and rubric dimensions. The same harness also runs stably on five backend LLMs from three model families, indicating that the harness generalizes across backends without backend-specific tuning.

In summary, our contributions are:
\textbf{(1)} We design OneDayAgent, a long-horizon harness that jointly addresses task decomposition, execution memory, and deliverable verification.
\textbf{(2)} Extensive experiments on AgentIF-OneDay show a new state of the art (0.821) and stable cross-backend generalization.
\textbf{(3)} We open-source the harness and trajectories to benefit the broader community.

\begin{figure*}[t!]
    \centering
    \includegraphics[width=1.0\linewidth]{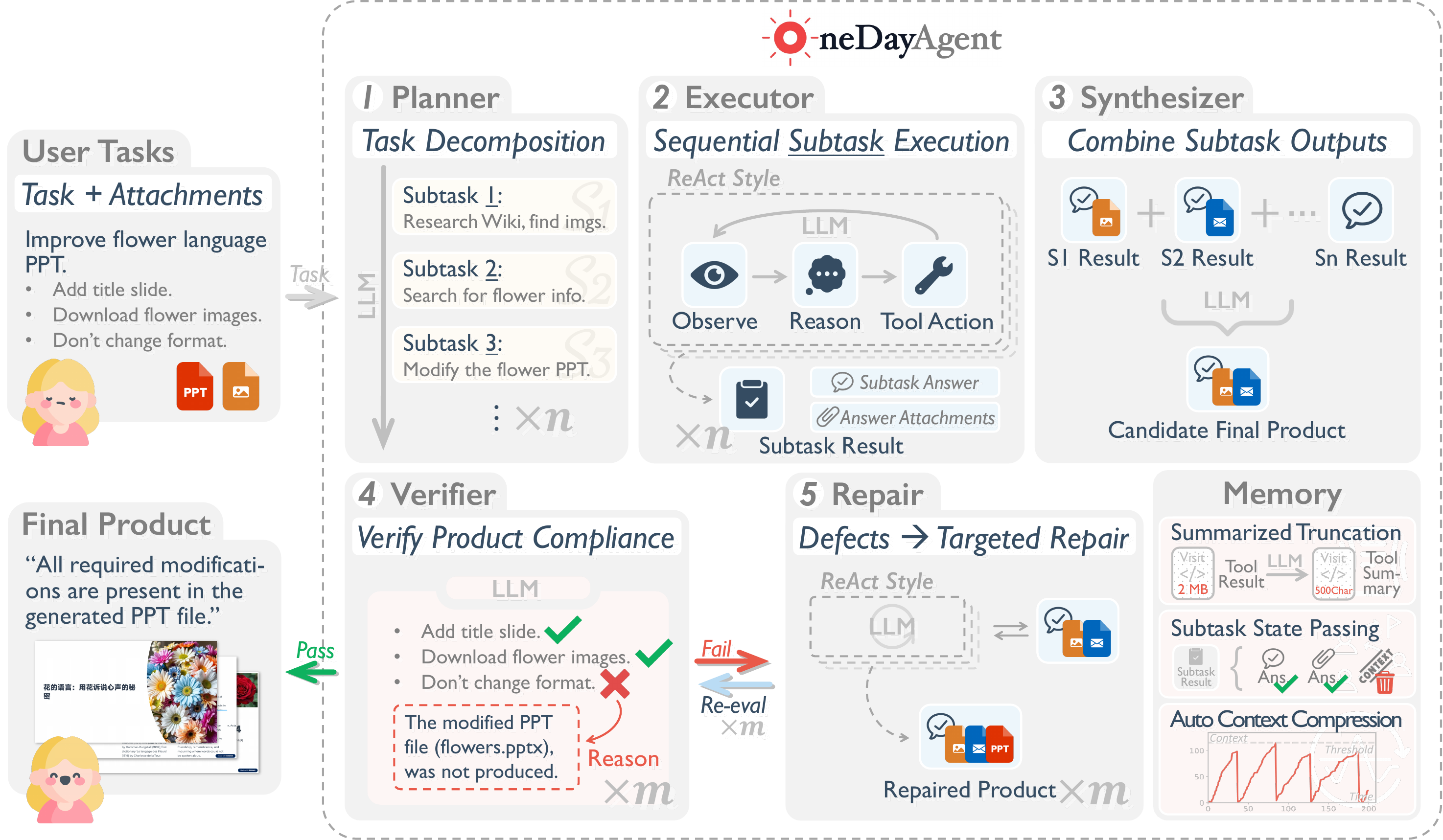}
    \caption{\textbf{Overview of OneDayAgent.}
    OneDayAgent uses a long-horizon harness that decomposes an everyday task, executes subtasks through environment-grounded tools, maintains execution memory, and performs global verification and repair before producing the final deliverable.}
    \label{fig:main_method}
\end{figure*}

\section{OneDayAgent}
\label{sec:method}

We target long-horizon everyday agency, where open-ended requests require multi-step progress over trajectories.
As shown in Figure~\ref{fig:intro}(a), these tasks have three characteristics.
First, they are long-horizon. The agent must preserve goals and constraints across many reasoning and action steps.
Second, they are cross-environment. Progress often requires moving between web pages, local files, code execution, generated artifacts, and external services.
Third, they are multimodal. Inputs and evidence may include text, documents, images, tables, and other attachments.
Together, these characteristics create three execution challenges. The agent may forget earlier constraints, lose or fail to pass intermediate state, and exceed the usable context budget before the final deliverable is complete.

To address these challenges, we design OneDayAgent as a long-horizon execution harness for open-ended everyday tasks.
It decomposes requests to make extended tasks tractable, verifies and repairs final deliverables to counter goal and constraint forgetting, maintains execution memory to preserve intermediate state under context pressure, and unifies tool and environment interaction for cross-environment and multimodal work.

\subsection{OneDayAgent Overview: From Task Intent to Deliverable}

OneDayAgent is a long-horizon harness that turns an everyday request into a managed execution process. A single uninterrupted ReAct~\cite{yao2023reactsynergizingreasoningacting} trajectory struggles to preserve goals and manage context over extended horizons, so OneDayAgent introduces explicit decomposition, memory, and verification stages into the execution loop. As shown in Figure~\ref{fig:main_method}, the workflow starts from a user request, together with any files, images, or other attachments that define the task context. The harness keeps this original request as the global intent and converts it into an ordered set of subtasks, giving the backend LLM a shorter local objective at each step while preserving the end-to-end deliverable as the target of the whole trajectory. Each subtask is then executed inside a ReAct loop where the backend LLM reasons about the current subtask, calls tools through the harness, observes environment feedback, and updates the working state. The tool interface spans the main environments needed by everyday tasks, as summarized in Table~\ref{tab:tool_environment}. Intermediate findings and produced artifacts are written into execution memory and the workspace, so later subtasks can build on earlier results without replaying the full interaction history. After all planned subtasks finish, OneDayAgent synthesizes the accumulated state and artifacts into a final deliverable. The harness then runs a global verification pass against the original request, the execution trace, and the produced output. If the verifier finds missing requirements or inconsistent artifacts, OneDayAgent enters a targeted repair loop and updates the deliverable before returning it. The overall workflow therefore follows a simple execution path, preserving task intent, decomposing long work, executing subtasks in real environments, maintaining execution memory, and verifying the final result before delivery.

\subsection{Harness Strategy: Structuring Long-Horizon Execution}

Having described the end-to-end workflow, we now unpack the harness into three core capabilities that address the failure modes of long-horizon everyday agency: task overload, goal drift, and state loss.
Implementation-level configuration values for these mechanisms are summarized in Appendix Table~\ref{tab:harness_configuration}.
The core runtime prompts that instantiate these stages are provided in Appendix~\ref{app:prompts}.

\begin{table}[t]
\centering
\small
\renewcommand{\arraystretch}{1.32}
\setlength{\tabcolsep}{2.5pt}
\newcolumntype{Y}{>{\arraybackslash}X}
\newcolumntype{L}[1]{>{\raggedright\arraybackslash}p{#1}}
\begin{tabularx}{0.8\columnwidth}{L{0.2\columnwidth} L{0.27\columnwidth} Y}
\toprule
\textbf{Group} & \textbf{Tools} & \textbf{Role} \\
\midrule
Web access & Search, visit & Retrieve web evidence \\
Academic search & Google Scholar, OpenAlex & Retrieve academic metadata \\
Computation & Python, command execution & Execute code and inspect outputs \\
File workspace & Read, write, edit files & Persist state and deliverables \\
Multimodal processing & Image analysis, generation & Interpret and create visual artifacts \\
\bottomrule
\end{tabularx}
\caption{\textbf{Tool and environment interface in OneDayAgent.}
The harness exposes heterogeneous environments through a small set of functional tool groups.
Implementation-level tool names are omitted here and listed in Appendix Table~\ref{tab:tool_environment_full}.}
\label{tab:tool_environment}
\end{table}

\subsubsection{Capability I: Task Decomposition}
\textbf{Task decomposition turns an overloaded long-horizon request into bounded executable units.}
OneDayAgent decomposes the original request into an ordered list of subtasks, corresponding to the planner in Panel 1 of Figure~\ref{fig:main_method}.
This follows the idea that long-horizon execution benefits from subgoal or hierarchical structure~\cite{wang2026subgoaldrivenframeworkimprovinglonghorizon,diao2026hipifhierarchicalplanninginformation}.
The design motivation is practical. Everyday requests often combine implicit requirements and artifact-level constraints, making a single uninterrupted executor trajectory easy to overload~\cite{zhou2024webarenarealisticwebenvironment}.
Decomposition gives the backend LLM a local objective at each step, while the harness keeps the original request as the global intent that every subtask must ultimately serve.
As shown in Panel 2, each subtask acts as an executable unit that can call tools, produce artifacts, and submit a compact answer.
The subtask boundary also becomes a context-saving interface.
Later subtasks inherit the accumulated task-level state, but not the full low-level ReAct trace.
After all subtasks finish, the synthesizer in Panel 3 combines the submitted subtask answers and attachments into a candidate final product.

\subsubsection{Capability II: Global Verification and Repair}
OneDayAgent treats final checking and repair as two tightly coupled stages that turn a candidate deliverable into a verified, task-aligned output.

\textbf{Global verification re-aligns the final deliverable with the original intent.}
Completing every subtask does not guarantee that the deliverable satisfies the original request.
Long-horizon execution can still lose early constraints, skip implicit requirements, or produce artifacts that are locally plausible but globally incomplete.
To catch these failures, OneDayAgent performs a global verification pass after synthesis, corresponding to Panel 4 of Figure~\ref{fig:main_method}.
The verifier checks the candidate final product against the original request, the submitted subtask answers, and the declared attachments, rather than judging only the final text response.
This design follows the feedback-and-revision pattern in LLM agents and generation systems~\cite{shinn2023reflexionlanguageagentsverbal,madaan2023selfrefineiterativerefinementselffeedback}, while making the check artifact-level and task-global.

\textbf{Targeted repair converts verification failures into localized execution updates.}
If verification finds a defect, OneDayAgent enters the ReAct-style repair stage shown in Panel 5 of Figure~\ref{fig:main_method}.
The repair step is targeted.
The harness uses the verifier's defect description to revise the missing or inconsistent part of the deliverable instead of restarting all subtasks.
This failure-to-fix style is aligned with recent work on verification-centric and repair-oriented agent systems~\cite{zhu2026marcodeepresearchunlockingefficient,mulian2026agentfixerfailuredetectionfix}.
After repair, the updated deliverable is re-evaluated, so verification acts as a final task-level guard rather than a passive scoring step.

\subsubsection{Capability III: Execution Memory}
OneDayAgent uses the memory block in Figure~\ref{fig:main_method} to make long-horizon execution state both compact and recoverable.
The goal is not to store every token, but to keep the information that later reasoning, tool use, and artifact construction actually depend on.
This is necessary because stateful long-horizon agent workloads require memory mechanisms that preserve task-relevant information without carrying every interaction token forward~\cite{omri2026agentmemorycharacterizationimplications,xu2026memgymlonghorizonmemoryenvironment}.

\textbf{Summarized truncation compresses high-volume noisy observations into reusable evidence.}
At the tool layer, raw observations from search, web visits, and local files can be much larger than the decision they support.
OneDayAgent therefore converts them into bounded evidence before they dominate the dialogue context.
Search outputs are kept as structured snippets, long pages are summarized with a bounded raw prefix, and file reads are reduced to modality-aware previews.
This keeps the executor grounded in external environments while filtering high-volume noisy observations, following the broader motivation of lightweight memory-augmented generation~\cite{fang2025lightmem}.

\textbf{Subtask state passing preserves task-level progress while discarding low-level traces.}
As noted in Capability I, subtask boundaries also define what state is carried forward.
OneDayAgent uses the submitted answer and declared result-file handles as compact checkpoints across environments and modalities.
This lets later subtasks reuse earlier files, images, search evidence, or generated artifacts without inheriting the full low-level ReAct trace.

\textbf{Automatic context compression keeps long-horizon execution inside the backend context budget.}
At the dialogue layer, OneDayAgent monitors the accumulated trajectory before continuing ReAct execution.
When the context exceeds a configurable fraction of the backend window, earlier interaction rounds are compressed into an LLM-generated technical summary while the system prompt, original user task, and recent action rounds remain available.
If execution approaches the hard limit, the harness falls back to deterministic emergency pruning of low-value history.
This makes memory an execution-control mechanism that preserves enough state to continue while preventing context pressure from becoming the bottleneck.

\subsubsection{Tool and Environment Interface}
OneDayAgent keeps tool and environment handling lightweight, exposing the resources needed for everyday tasks without making them the center of the method.

\textbf{Unified tool use exposes heterogeneous environments through one ReAct action space.}
OneDayAgent wraps web access, academic search, computation, file operations, and multimodal processing as unified tool actions that can be called directly inside the same shared observe-reason-act loop.
Table~\ref{tab:tool_environment} summarizes the main tool groups, and Appendix Table~\ref{tab:tool_environment_full} lists the concrete tool interface.

\textbf{Workspace artifacts make environment interactions persistent across the workflow.}
Tool calls can return text observations, but they can also create or modify files, images, code outputs, and other artifacts.
The harness keeps these artifacts in the task workspace and result-file state, so later subtasks, synthesis, verification, and repair can refer to concrete environment outputs rather than relying only on transient dialogue history.

\section{Experiments}

We evaluate OneDayAgent from five perspectives.
We first report main performance on AgentIF-OneDay, then study ablations, execution behavior, backend transferability, and a concrete case study.

\begin{table*}[t]
    \centering
    \small
    \renewcommand{\arraystretch}{1.35}
    \setlength{\tabcolsep}{2.5pt}

    \newcommand{\best}[1]{\textbf{#1}}
    \newcommand{\na}{--}
    \newcolumntype{Y}{>{\centering\arraybackslash}X}
    \newcolumntype{C}[1]{>{\centering\arraybackslash}p{#1}}
    \definecolor{onedayyellow}{HTML}{F4D35E}

    \begin{tabularx}{\textwidth}{p{0.195\textwidth} | YYY | YYY | YYY | C{0.062\textwidth} C{0.062\textwidth} | C{0.074\textwidth} C{0.049\textwidth}}
        \toprule
        \multicolumn{14}{c}{AgentIF-OneDay Benchmark\rlap{\hspace{0.35em}\smash{\raisebox{-0.15em}{\includegraphics[height=1.2em]{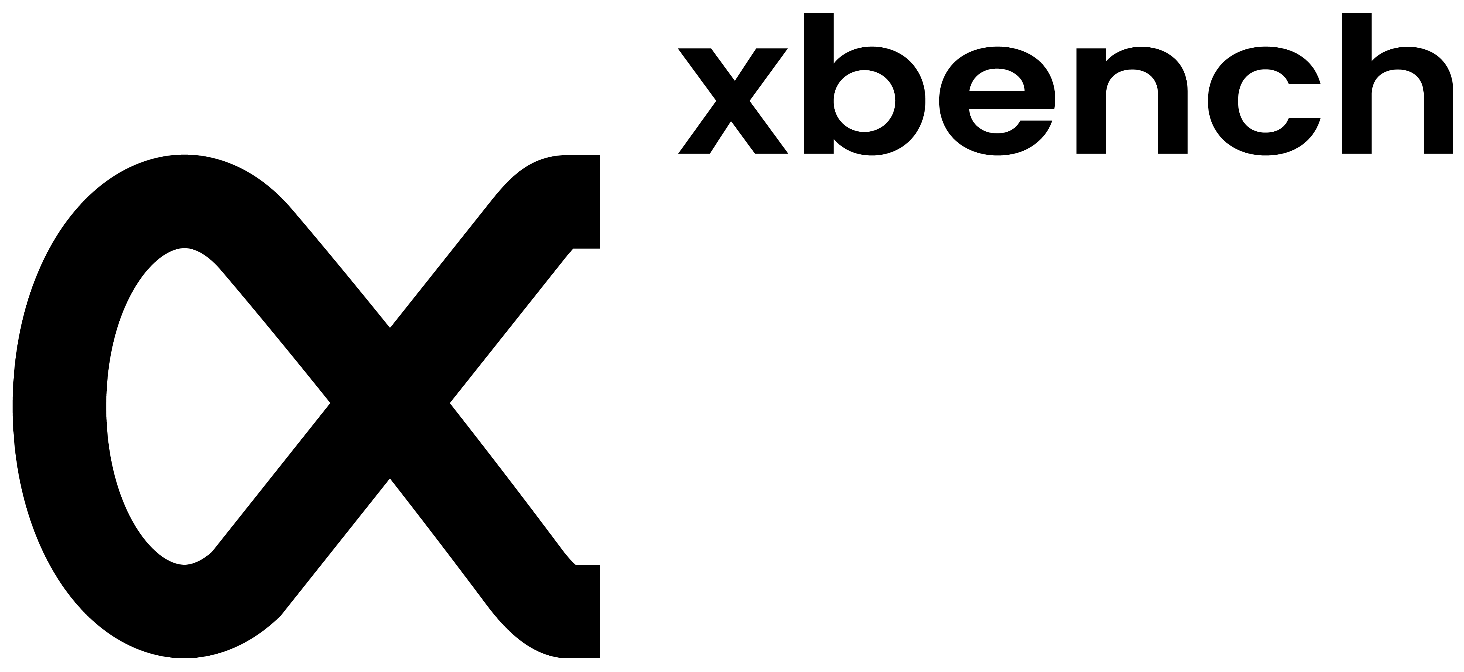}}}}} \\
        \midrule
        \multicolumn{1}{r|}{\textbf{Metric [0,1]} \textbf{$\rightarrow$}} &
        \multicolumn{3}{c|}{\textit{Task Type}} &
        \multicolumn{3}{c|}{\textit{Domain}} &
        \multicolumn{3}{c|}{\textit{Rubric}} &
        \multicolumn{2}{c|}{\textit{Input Attachment}} &
        \multicolumn{2}{c}{\textit{\textbf{Overall}}} \\
        \cmidrule(lr){2-4}
        \cmidrule(lr){5-7}
        \cmidrule(lr){8-10}
        \cmidrule(lr){11-12}
        \cmidrule(lr){13-14}
        \textbf{Methods / Backends} $\downarrow$ & OWE & LII & IR
        & Work & Life & Study
        & Inst. & Fact. & Logic
        & w/o & w/
        & \mbox{Latency(s)} & \textbf{Score} \\
        \midrule

        \rowcolor{gray!10}
        \multicolumn{14}{c}{\textbf{Baseline Agents}} \\
        \midrule

        Minimax-Agent\textsuperscript{\dag}
        & 0.525 & 0.510 & 0.717
        & \na & \na & \na
        & 0.709 & 0.623 & 0.755
        & 0.502 & 0.603
        & 1416.2 & \cellcolor{gray!6}0.562 \\

        ChatGPT-Agent\textsuperscript{\dag}
        & 0.606 & 0.613 & 0.689
        & 0.722 & 0.697 & 0.593
        & 0.739 & 0.687 & 0.673
        & 0.566 & 0.666
        & 665.1 & \cellcolor{gray!6}0.626 \\

        Genspark\textsuperscript{\dag}
        & 0.577 & 0.719 & 0.681
        & 0.719 & 0.679 & 0.712
        & 0.766 & 0.663 & 0.720
        & 0.551 & 0.691
        & 484.1 & \cellcolor{gray!6}0.635 \\

        Manus\textsuperscript{\dag}
        & 0.661 & 0.610 & 0.646
        & 0.703 & 0.734 & 0.644
        & 0.762 & 0.731 & 0.693
        & 0.644 & 0.646
        & 500.0 & \cellcolor{gray!6}0.645 \\

        Codex(GPT-5.5 medium)
        & 0.682 & 0.648 & 0.638
        & 0.740 & 0.584 & 0.529
        & 0.651 & 0.672 & 0.665
        & 0.613 & 0.699
        & \best{325.5} & \cellcolor{gray!6}0.664 \\

        AutoClaw\textsuperscript{\dag}
        & \na & \na & \na
        & \na & \na & \na
        & \na & \na & \na
        & \na & \na
        & 523.0 & \cellcolor{gray!6}0.799 \\

        \midrule
        \rowcolor{onedayyellow!12}
        \multicolumn{14}{c}{\textbf{\vphantom{Baseline Agents}\smash{\raisebox{-0.55em}{\includegraphics[height=1.85em]{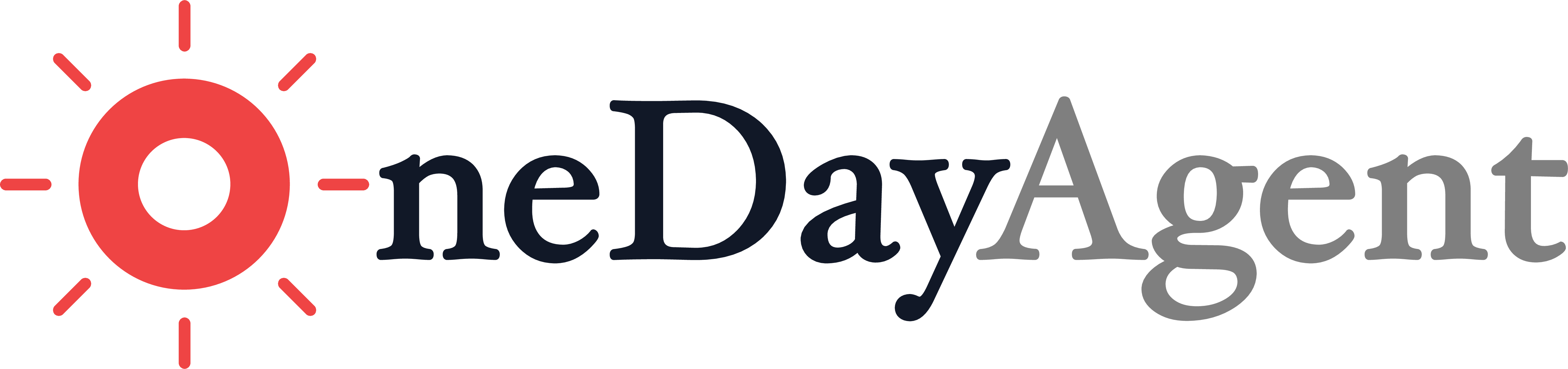}}} (OneDayAgent Backend Variants)}} \\
        \midrule

        \rowcolor{onedayyellow!4}
        Qwen3.6-27B
        & 0.606 & 0.649 & 0.589
        & 0.628 & 0.725 & 0.491
        & 0.543 & 0.617 & 0.769
        & 0.611 & 0.615
        & 1280.5 & \cellcolor{onedayyellow!15}0.613 \\

        \rowcolor{onedayyellow!4}
        Qwen3.5-9B
        & 0.654 & 0.608 & 0.565
        & 0.674 & 0.615 & 0.500
        & 0.607 & 0.611 & 0.724
        & 0.625 & 0.623
        & 1895.2 & \cellcolor{onedayyellow!15}0.624 \\

        \rowcolor{onedayyellow!4}
        Qwen3.5-397B-A17B
        & 0.763 & 0.623 & 0.666
        & 0.728 & 0.682 & 0.674
        & 0.671 & 0.718 & 0.750
        & 0.722 & 0.698
        & 964.5 & \cellcolor{onedayyellow!15}0.708 \\

        \rowcolor{onedayyellow!4}
        \mbox{Gemini-3.1-Pro-Preview}
        & 0.714 & 0.806 & 0.743
        & 0.774 & 0.730 & 0.674
        & 0.783 & 0.714 & 0.776
        & 0.651 & 0.806
        & 1281.6 & \cellcolor{onedayyellow!15}0.743 \\

        \rowcolor{onedayyellow!4}
        GLM-5.2
        & \best{0.818} & \best{0.821} & \best{0.829}
        & \best{0.855} & \best{0.823} & \best{0.731}
        & \best{0.784} & \best{0.835} & \best{0.846}
        & \best{0.782} & \best{0.847}
        & 3216.8 & \cellcolor{onedayyellow!15}\best{0.821} \\

        \bottomrule
    \end{tabularx}
    \caption{\textbf{Main Results on AgentIF-OneDay.}
    We compare OneDayAgent with general-purpose agents across task types, domains, rubric dimensions, input-attachment settings, latency, and overall score $[0,1]$. OWE, LII, and IR denote Open Workflow Execution, Latent Instruction Inference, and Iterative Refinement, while Inst., Fact., and Logic denote Instruction Following, Factuality, and Logic/Functionality.
    The w/ and w/o columns report scores with and without input attachments.
    \textsuperscript{\dag} marks official AgentIF-OneDay results evaluated with Gemini-3-Pro-Preview as the LLM-as-judge and validated against human annotations.
    \textbf{With the GLM-5.2 backend, OneDayAgent achieves the strongest overall score of 0.821 and leads all score dimensions.}}
    \label{tab:main_results}
\end{table*}

\subsection{Experimental Setup}

\subsubsection{AgentIF-OneDay Benchmark}

We evaluate OneDayAgent on AgentIF-OneDay~\cite{chen2026agentifonedaytasklevelinstructionfollowingbenchmark}, a task-level benchmark for instruction following in general-purpose agents across daily-life scenarios.
Unlike short question-answering benchmarks, AgentIF-OneDay requires agents to complete realistic daily tasks with attachments, multimodal evidence, and concrete deliverables.
The benchmark contains 104 tasks and 767 instance-level scoring points, covering work, study, and life scenarios.

AgentIF-OneDay organizes tasks into three user-interaction patterns.
Open Workflow Execution (OWE) tests whether the agent can follow an explicit multi-step procedure without dropping key constraints; Latent Instruction Inference (LII) requires the agent to infer implicit rules from provided materials and apply them faithfully to a new task; and Iterative Refinement (IR) evaluates whether the agent can modify or extend an existing artifact while maintaining consistent state.
Evaluation uses binary instance-level rubrics with bonus and penalty criteria, where satisfied positive criteria add points, triggered penalty criteria subtract points, and the resulting task score is clipped and normalized to $[0,1]$ before averaging across tasks.

\subsubsection{Implementation and Evaluation Setup}

We use GLM-5.2~\cite{glm5team2026glm5vibecodingagentic} as the backend for the main results, ablation, and behavior analysis.
For the backend analysis, we evaluate four additional LLMs: Gemini-3.1-Pro-Preview~\cite{deepmind2026gemini31pro} (June 2026), Qwen3.5-397B-A17B and Qwen3.5-9B~\cite{qwen3.5}, and Qwen3.6-27B~\cite{qwen3.6-27b}.
All backend runs use the same OneDayAgent harness with temperature 1.0, top-$p$ 0.95, 128K max tokens, 200 ReAct iterations, 7200-second timeout, up to 6 subtasks, and context compression at 0.9$\times$ budget. Full configuration is in Appendix~\ref{tab:harness_configuration}.

For comparison, we report general-purpose agent baselines from the official AgentIF-OneDay release~\cite{chen2026agentifonedaytasklevelinstructionfollowingbenchmark} together with an additional Codex run using GPT-5.5 medium~\cite{openai2026gpt55systemcard}.

We report normalized AgentIF-OneDay scores in $[0,1]$, mean latency, and aggregate harness-behavior metrics.
For scoring, we use the official AgentIF-OneDay LLM-as-judge framework.
The official release used Gemini-3-Pro-Preview~\cite{deepmind2026gemini3pro} as the judge, but that model was no longer available in our evaluation environment (June 2026). 
We therefore use Gemini-3.1-Pro-Preview~\cite{deepmind2026gemini31pro} with the same parameter settings.
A paired comparison on the same run shows Gemini-3.1-Pro-Preview scores 3.12 percentage points lower than Gemini-3-Pro-Preview, making our scores conservative relative to baselines (Appendix~\ref{app:judge_comparison}).

\subsection{Main Results}

\textbf{OneDayAgent sets a new state of the art on AgentIF-OneDay.}
Table~\ref{tab:main_results} shows that OneDayAgent with the GLM-5.2 backend achieves the best overall score, 0.821, outperforming both official general-purpose agent baselines and our additional Codex run.
The gain is not limited to a single slice of the benchmark, since OneDayAgent leads across all task types, domains, rubric dimensions, and input-attachment settings.
This indicates that the harness improves broad task-level instruction following rather than only optimizing one narrow evaluation category.

\subsection{Ablation Study}

\begin{table*}[t]
\centering
\small
\renewcommand{\arraystretch}{1.25}
\setlength{\tabcolsep}{2.5pt}
\newcolumntype{Y}{>{\centering\arraybackslash}X}
\newcolumntype{C}[1]{>{\centering\arraybackslash}p{#1}}
\begin{tabularx}{\textwidth}{p{0.078\textwidth} Y | C{0.073\textwidth} C{0.08\textwidth} | C{0.086\textwidth} C{0.052\textwidth} C{0.063\textwidth} | C{0.093\textwidth} C{0.057\textwidth} C{0.043\textwidth} C{0.08\textwidth}}
\toprule
\multirow{2}{*}{\textbf{Variant}} & \multirow{2}{*}{\textbf{Removed Module}}
& \multicolumn{2}{c|}{\textit{Performance}}
& \multicolumn{3}{c|}{\textit{Behavior}}
& \multicolumn{4}{c}{\textit{Efficiency / Difference}} \\
\cmidrule(lr){3-4}\cmidrule(lr){5-7}\cmidrule(lr){8-11}
& & \mbox{Overall} & \mbox{$\Delta$ Direct} & \mbox{Lat. (min)} & \mbox{Tools} & \mbox{Repair} & \mbox{Score/Lat.} & \mbox{Perfect} & \mbox{Zero} & \mbox{Better Full} \\
\midrule
DIRECT & Decompose + Verify & 0.771 & -- & \textbf{27.6} & 28.4 & -- & \textbf{2.80} & 52 & 9 & 12 \\
DECOMP & Verify & 0.804 & +3.3 pp & 38.1 & 45.7 & -- & 2.11 & 54 & 3 & 13 \\
VERIFY & Decompose & 0.804 & +3.3 pp & 29.7 & 29.3 & 3.9\% & 2.71 & 51 & \textbf{2} & \textbf{17} \\
FULL & None & \textbf{0.821} & \textbf{+5.0 pp} & 53.6 & 51.6 & 8.6\% & 1.53 & \textbf{58} & 5 & -- \\
\bottomrule
\end{tabularx}
\caption{\textbf{Ablation results for decomposition and verification modules.}
All variants use GLM-5.2 and are evaluated on the same 104 AgentIF-OneDay tasks.
DIRECT disables both decomposition and verification; DECOMP keeps only decomposition; VERIFY keeps only verification; FULL enables both modules.
Score / Lat. reports percentage-point score divided by mean latency in minutes.
``Better Full'' counts tasks where the variant scores higher than FULL.}
\label{tab:ablation_main}
\end{table*}

This ablation asks whether harness modules improve success and whether extra cost is justified by the gain.

\textbf{Harness modules improve task success.}
Table~\ref{tab:ablation_main} isolates decomposition and verification in a 2$\times$2 ablation with the same GLM-5.2 backend.
Execution memory remains enabled in all variants, as disabling it causes context overflow or state loss that prevents task completion.
Starting from DIRECT, which disables both modules, decomposition alone improves the overall score from 0.771 to 0.8039, and verification alone reaches a nearly identical 0.8044.
Enabling both modules gives the best score, 0.821, showing that both mechanisms contribute to final task success.
The combined gain is smaller than the sum of the two isolated gains, suggesting that the modules partly recover overlapping failure cases.

\textbf{Module gains come with very different costs.}
The same table shows a large cost asymmetry, with VERIFY adding only 2.2 minutes over DIRECT while matching the score of DECOMP, whereas DECOMP adds 10.6 minutes and increases tool calls by roughly 60\%.
The efficiency columns make this tradeoff explicit, as VERIFY remains close to DIRECT in latency while reaching the same score as DECOMP, whereas FULL obtains the highest score but the lowest score-per-latency ratio.
Thus, the full harness is best when score is the primary objective, but verification-only is the strongest cost-effective point.

\textbf{Always-on module composition is not uniformly optimal.}
FULL produces the largest number of perfect tasks, but simpler variants still outperform FULL on a non-trivial subset.
In Table~\ref{tab:ablation_main}, VERIFY scores higher than FULL on 17 tasks, DECOMP on 13, and DIRECT on 12.
This pattern suggests that enabling every harness module raises the ceiling, but the best configuration depends on whether the priority is maximum score or lower execution cost.

\begin{figure*}[t]
    \centering
    \includegraphics[width=\textwidth]{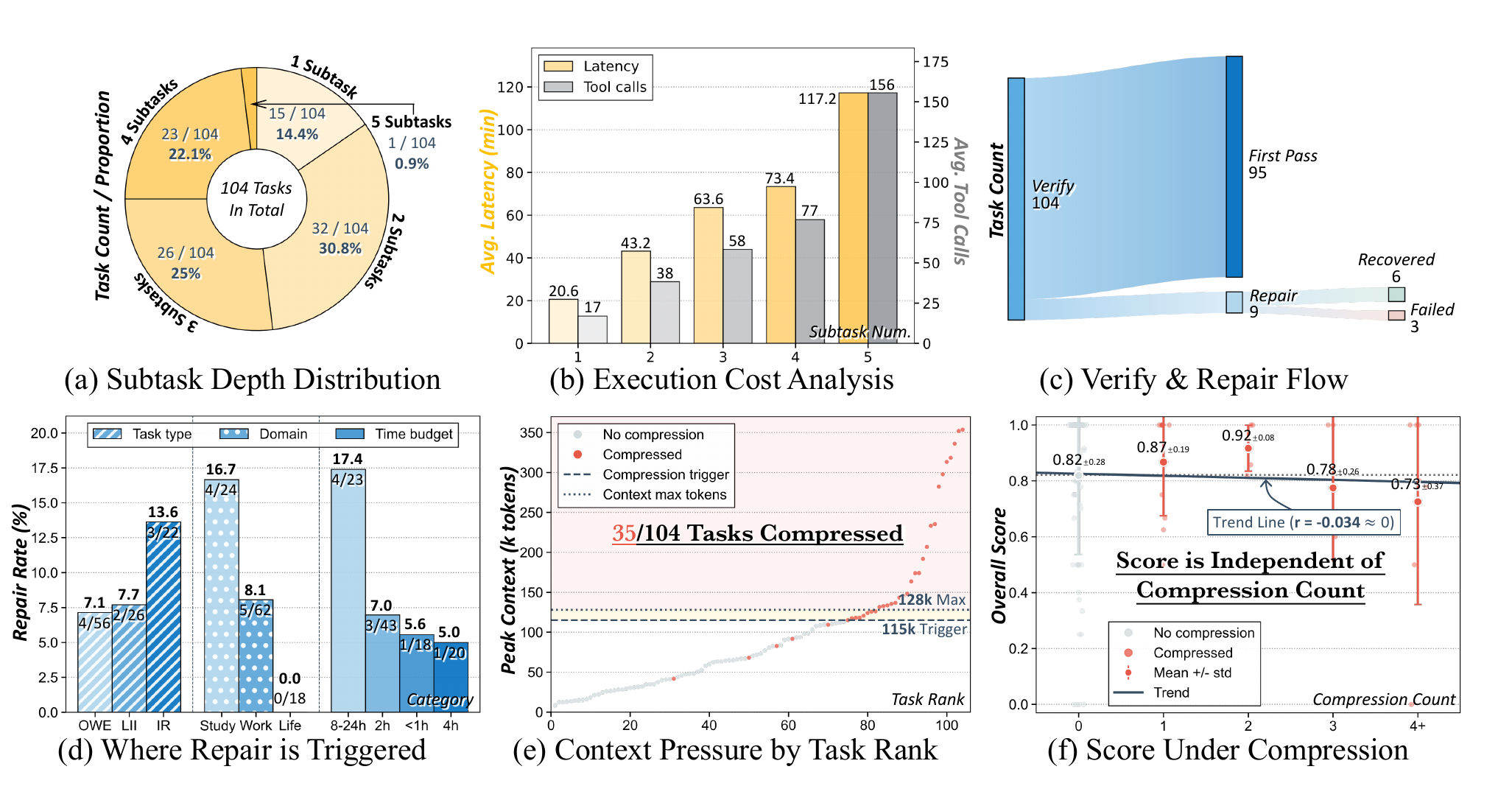}
    \caption{\textbf{Execution behavior of OneDayAgent.}
    OneDayAgent turns long-horizon tasks into a managed execution process through decomposition, verify/repair, and context management.}
    \label{fig:backend_execution_analysis}
\end{figure*}

\subsection{Execution Behavior}

We next analyze how OneDayAgent manages long-horizon execution pressure inside the GLM-5.2 run.

\textbf{Decomposition turns open-ended work into structured execution.}
Figure~\ref{fig:backend_execution_analysis}(a) shows that OneDayAgent rarely treats AgentIF-OneDay tasks as a single uninterrupted trajectory, with most tasks decomposed into two to four subtasks and only 16 of 104 tasks executed as one subtask.
Figure~\ref{fig:backend_execution_analysis}(b) further shows that deeper decompositions correspond to higher execution cost, increasing from 20.6 minutes and 17 tool calls for one-subtask tasks to 117.2 minutes and 156 tool calls for five-subtask tasks.
Decomposition depth therefore correlates with tractability and complexity, as the harness converts open-ended work into bounded executable units, while harder tasks still require more time and tool interaction.

\textbf{Verification and repair make delivery risk observable and recoverable.}
Figure~\ref{fig:backend_execution_analysis}(c) shows that 95 of 104 tasks pass verification on the first attempt, while 9 enter repair; among those repaired tasks, 6 are recovered and 3 still fail.
The repair distribution in Figure~\ref{fig:backend_execution_analysis}(d) indicates that repair is concentrated in harder settings, especially IR tasks, study-domain tasks, and long time-budget tasks.
This supports the role of verify/repair as a delivery-risk mechanism rather than a generic score booster, because it exposes residual defects after execution and recovers some, but not all, difficult cases.

\textbf{Context management keeps long trajectories feasible.}
Figure~\ref{fig:backend_execution_analysis}(e) shows substantial context pressure, with 35 of 104 tasks triggering compression and the highest-pressure task accumulating roughly 350K context tokens across compression rounds.
At the same time, Figure~\ref{fig:backend_execution_analysis}(f) shows no systematic score degradation as compression count increases, with a near-zero correlation between compression count and score.
Thus, context management is associated with stable task quality under context pressure, though causal isolation is left to future work.

\subsection{Backend Analysis}

This section tests whether OneDayAgent is a backend-specific system or a transferable harness, and whether backend differences can be reduced to model scale alone.

\begin{table}[t]
\centering
\small
\renewcommand{\arraystretch}{1.25}
\setlength{\tabcolsep}{4pt}
\begin{tabular}{l l c c c c}
\toprule
\textbf{Backend} & \textbf{Family / Vendor} & \textbf{Scale} & \textbf{Success} & \textbf{Overall} & \textbf{Latency (s)} \\
\midrule
GLM-5.2 & GLM / Zhipu & 744B & 104/104 & \textbf{0.821} & 3216.8 \\
Gemini-3.1-Pro-Preview & Gemini / Google & undisclosed & 104/104 & 0.743 & 1281.6 \\
Qwen3.5-397B-A17B & Qwen / Alibaba & 397B-A17B & 104/104 & 0.708 & 964.5 \\
Qwen3.6-27B & Qwen / Alibaba & 27B & 104/104 & 0.613 & 1280.5 \\
Qwen3.5-9B & Qwen / Alibaba & 9B & 104/104 & 0.624 & 1895.2 \\
\bottomrule
\end{tabular}
\caption{\textbf{Backend coverage and performance under the same OneDayAgent harness.}
All backends are evaluated on the full 104-task AgentIF-OneDay suite.
Reported scale is shown when publicly available or encoded in the model name; Gemini-3.1-Pro-Preview scale is not publicly disclosed.}
\label{tab:backend_transferability}
\end{table}

\begin{figure*}[t]
    \centering
    \includegraphics[width=\textwidth]{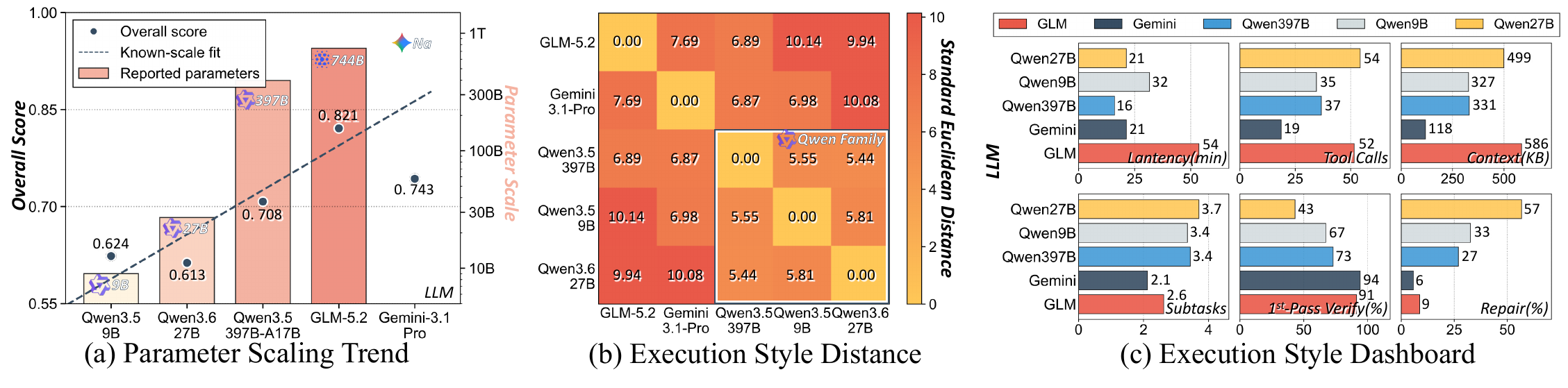}
    \caption{\textbf{Backend scaling and execution-style interaction.}
    Under the same OneDayAgent harness, backend performance shows a weak parameter-scaling trend rather than a strict scaling law, while backend-specific behavior appears in execution-style distance and operational profiles.}
    \label{fig:backend_scaling_analysis}
\end{figure*}

\textbf{The same OneDayAgent harness remains effective across heterogeneous backend LLMs.}
The backend-variant rows in Table~\ref{tab:main_results} provide the first evidence for transferability.
Without changing the OneDayAgent harness, all tested backends obtain non-trivial scores not only in overall performance, but also across task type, domain, rubric, and input-attachment slices.
If decomposition, tool use, context management, or verify/repair were tightly coupled to one backend, we would expect incomplete runs or collapse on particular benchmark slices.
Table~\ref{tab:backend_transferability} makes the coverage point explicit by listing five backend LLMs from different model families, vendors, and reported scales under the same 104-task AgentIF-OneDay suite.
Overall scores range from 0.613 to 0.821 across GLM/Zhipu, Gemini/Google, and Qwen/Alibaba backends.
Thus, the same harness transfers across model families and vendors, although final quality and latency vary substantially.

\textbf{Backend performance shows a parameter-scaling trend, but not a strict scaling law.}
Figure~\ref{fig:backend_scaling_analysis}(a) relates overall score to model scale.
Within disclosed-scale models, larger backends tend to perform better, visibly from Qwen3.5-9B (0.624) to Qwen3.5-397B-A17B (0.708) and GLM-5.2 (0.821).
However, the ordering is not monotonic, as Qwen3.6-27B does not dominate Qwen3.5-9B, and Gemini-3.1-Pro-Preview is widely believed to be larger than 1T, but still reaches only the second-best overall score.
This matches recent agentic-evaluation findings that conventional model rankings and scale alone do not fully predict tool-using agent performance~\cite{roig2025standardenterpriserelevantagentic,qi2025agentifbenchmarkinginstructionfollowing}.
Thus, parameter count is a useful axis for interpreting backend performance, but it is insufficient as a strict law for agentic long-horizon execution.

\textbf{Backend differences appear as execution-style differences under the same harness.}
The distance and execution-profile panels in Figure~\ref{fig:backend_scaling_analysis}(b,c) show that backend choice changes how the harness is used, not only the score.
GLM-5.2 obtains the strongest score but uses a high-cost profile, averaging 53.6 minutes, 51.6 tool calls, and 585.7 KB context per task.
Gemini-3.1-Pro-Preview follows a leaner profile with 21.4 minutes, 18.7 tool calls, and 118.1 KB context, while Qwen3.6-27B triggers the highest repair rate, 56.7\%.
The heatmap also shows weak family proximity among Qwen variants, but not enough to explain all behavior, with Qwen3.5-397B and Qwen3.6-27B as the closest pair while Qwen models still differ substantially in repair and first-pass verification.
These differences indicate that OneDayAgent is transferable as a harness, while backend LLMs induce distinct execution styles.

\subsection{Case Study}

\begin{figure*}[t]
    \centering
    \includegraphics[width=\textwidth]{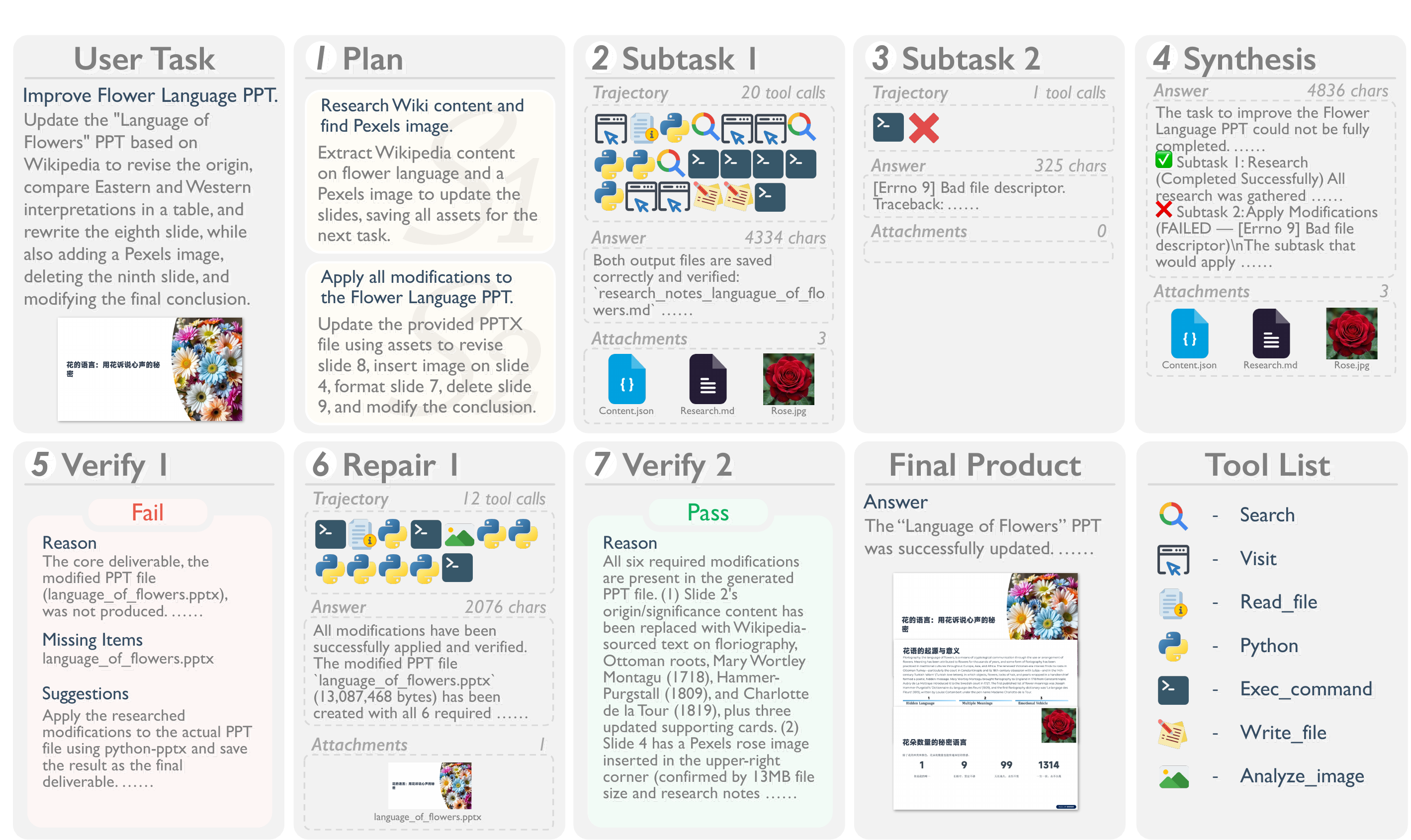}
    \caption{\textbf{Case study of a PPT-editing task.}
    The trajectory illustrates how OneDayAgent decomposes a multimodal editing request, exposes an incomplete subtask during synthesis, and uses verification-guided repair to produce the missing deliverable.}
    \label{fig:case_study}
\end{figure*}

Figure~\ref{fig:case_study} shows a representative ``Language of Flowers'' PPT-editing task.
The user asks the agent to revise the slide content using Wikipedia, compare Eastern and Western interpretations, insert a Pexels image, delete one slide, and update the final slide conclusion.

OneDayAgent decomposes the task into a research subtask and PPT modification subtask.
The first subtask collects needed text and image assets, while the second fails with a file-descriptor error.
During synthesis, the agent reports the failed subtask instead of marking the whole task as complete.
The verifier then identifies the missing PPT file and suggests applying the collected modifications to the actual deck.
The repair stage generates the missing presentation, and the second verification pass confirms that the requested slide edits and image insertion are present.

\section{Related Work}

\textbf{General-purpose agents for everyday digital tasks.}
LLM agents extend language models from text generation to problem solving, where the model reasons about intent, invokes tools, and reacts to environment feedback~\cite{luo2025largelanguagemodelagent,Plaat_2025,v2026agenticartificialintelligenceai,hu2026agentictooluselarge,zhang2025generalizabilitylargelanguagemodelbased}.
Methodologically, ReAct~\cite{yao2023reactsynergizingreasoningacting} grounds this view in interleaved reasoning and acting, Reflexion~\cite{shinn2023reflexionlanguageagentsverbal} and Self-Refine~\cite{madaan2023selfrefineiterativerefinementselffeedback} add feedback-based revision, and AutoGen~\cite{wu2023autogenenablingnextgenllm} organizes multi-agent workflows.
Concrete agent systems have also emerged across digital settings: WebSailor-V2~\cite{li2025websailorv2bridgingchasmproprietary} targets web information seeking, Agent S2~\cite{agashe2025agents2compositionalgeneralistspecialist} studies computer-use agents, AlphaEvolve~\cite{novikov2025alphaevolvecodingagentscientific}, ContextCov~\cite{sharma2026contextcovderivingenforcingexecutable}, and SEMAG~\cite{peng2026semagselfevolutionarymultiagentcode} develop coding and software-engineering agents, WebResearcher~\cite{qiao2025webresearcherunleashingunboundedreasoning}, Marco DeepResearch~\cite{zhu2026marcodeepresearchunlockingefficient}, and MiroThinker~\cite{miromindteam2026mirothinker17h1heavyduty} focus on deep research agents, ForeAgent~\cite{zheng2026predictexecutingmachinelearning} studies machine-learning research execution, and AgentRL~\cite{zhang2025agentrlscalingagenticreinforcement} and Temp-R1~\cite{gong2026tempr1unifiedautonomousagent} study training-based agent improvement.
In digital settings, agent research is also organized around concrete evaluation environments: WebArena~\cite{zhou2024webarenarealisticwebenvironment} and WebChoreArena~\cite{miyai2025webchorearenaevaluatingwebbrowsing} focus on web interaction, OSWorld2.0~\cite{yuan2026osworld20benchmarkingcomputeruse}, WindowsWorld~\cite{li2026windowsworldprocesscentricbenchmarkautonomous}, and MobileWorld~\cite{kong2025mobileworldbenchmarkingautonomousmobile} evaluate desktop or mobile software agents, SWE-Bench Pro~\cite{deng2025swebenchproaiagents} and DeepSWE~\cite{datacurve2026deepswev11} study software-engineering agents, and BrowseComp~\cite{wei2025browsecompsimplechallengingbenchmark}, DeepResearch Bench~\cite{du2025deepresearchbenchcomprehensivebenchmark}, $\tau^2$-Bench~\cite{barres2025tau2benchevaluatingconversationalagents}, and recent expert-level academic benchmarks~\cite{expert2026benchmarkacademicquestions} evaluate browsing, research, or conversational assistance. Early LLM evaluation~\cite{chen2021evaluatinglargelanguagemodels} and task-completion benchmarks~\cite{guo2026questionansweringtaskcompletion} laid the groundwork for these settings.
OneDayAgent instead follows AgentIF-OneDay~\cite{chen2026agentifonedaytasklevelinstructionfollowingbenchmark} in targeting everyday requests across work, life, and study.

\textbf{Harnesses for long-horizon agent execution.}
For long-horizon tasks, reliability depends not only on the backend model but also on the execution harness around it.
This setting is increasingly reflected in long-horizon and cross-environment benchmarks such as Tool Decathlon~\cite{li2026tooldecathlonbenchmarkinglanguage}, LifeSim~\cite{duan2026lifesimlonghorizonuserlife}, AgencyBench~\cite{li2026agencybenchbenchmarkingfrontiersautonomous}, OdysseyArena~\cite{yan2026odysseyarenabenchmarkinglargelanguage}, WeaveBench~\cite{li2026weavebenchlonghorizonrealworldbenchmark}, Workspace-Bench~\cite{tang2026workspacebench10benchmarkingai}, and Terminal-Bench~\cite{merrill2026terminalbenchbenchmarkingagentshard}, with additional benchmarks targeting verifiable constraints~\cite{zhang2026deepplanningbenchmarkinglonghorizonagentic}, interactive real-world workflows~\cite{shen2026tripbenchbenchmarklonghorizoninteractive}, large-scale tool ecosystems~\cite{liu2026planbenchxlevaluatinglonghorizonplanning}, multilingual settings~\cite{li2026polyworkbenchbenchmarkingmultilinguallonghorizon}, and strategic decision-making~\cite{zhang2026retailbenchevaluatinglonghorizonautonomous}, where success depends on coordinating information, intermediate artifacts, and final deliverables.
Recent work treats harnesses as runtime and state substrates for software agents~\cite{zhong2026aiharnessengineeringruntime,pysklo2026agentdiffbenchmarkingllmagents}.
One common line improves planning through search-style reasoning scaffolds, subgoal decomposition, or hierarchical planning for long-horizon agents~\cite{yao2023treethoughtsdeliberateproblem,wang2026subgoaldrivenframeworkimprovinglonghorizon,diao2026hipifhierarchicalplanninginformation}.
Another line manages long contexts and persistent state, including context folding~\cite{sun2025scalinglonghorizonllmagent,chen2026dynamiclongcontextreasoning}, agent memory characterization~\cite{omri2026agentmemorycharacterizationimplications}, long-horizon memory environments~\cite{xu2026memgymlonghorizonmemoryenvironment}, memory-augmented structures~\cite{xu2025amemagenticmemoryllm,fang2025lightmem}, broader analyses of memory mechanisms in foundation models~\cite{huang2026rethinkingmemorymechanismsfoundation}, learned context curation~\cite{zhang2026memoryactionautonomouscontext}, and hierarchical memory benchmarked against RAG and summarization~\cite{raheem2026contextcollapse}.
InfiAgent~\cite{yu2026infiagentinfinitehorizonframeworkgeneralpurpose} pursues the same long-horizon goal through file-centric state externalization with a strictly bounded reasoning context, whereas OneDayAgent keeps state compact via subtask checkpointing and context compression while additionally decomposing work and verifying the final deliverable.
Verification and repair form a third line, with recent work emphasizing failure diagnosis and repair recommendations~\cite{mulian2026agentfixerfailuredetectionfix}, trajectory-level failure taxonomies~\cite{wang2026longhorizontaskmiragediagnosing}, and adversarial robustness against long-horizon attacks~\cite{jiang2026agentlabbenchmarkingllmagents}.
Tool and workflow orchestration is also central, as shown by real MCP tool-use evaluation~\cite{bandi2026mcpatlaslargescalebenchmarktooluse}.
OneDayAgent builds on this line with a harness that plans subtasks, executes them sequentially, passes memory and state through intermediate results, manages context, and verifies or repairs the final output.
This framing lets us study the harness as a transferable layer, while also measuring how different backend LLMs change execution behavior under the same workflow.

\section{Conclusion}

We presented OneDayAgent, a long-horizon harness for autonomous agents.
OneDayAgent turns open-ended requests into a managed execution process that decomposes tasks into bounded subtasks, maintains execution memory under context pressure, and verifies and repairs the final deliverable.
On AgentIF-OneDay, the GLM-5.2 backend achieves the best overall score of 0.821, and the same unchanged harness runs stably across five backend LLMs from three model families.
Our findings suggest two takeaways for long-horizon agent research.
First, a single harness can manage decomposition, memory, and verification jointly, without being tailored to one backend.
Second, cross-backend transfer is not silent, as different models induce distinct execution styles under the same workflow, including latency, tool-call volume, and repair rate.
These findings are specific to AgentIF-OneDay; broader generalization requires validation on additional benchmarks. The current implementation runs without workspace isolation; see Appendix~\ref{app:security} for security considerations.

\bibliographystyle{unsrtnat}
\bibliography{reference}

\clearpage
\appendix
\section*{Appendix}
\addcontentsline{toc}{section}{Appendix}

\section{Tool Interface Details}
\label{app:tool-interface-details}

Table~\ref{tab:tool_environment_full} expands the tool groups used in the main text into the concrete runtime interfaces exposed to OneDayAgent.
We list each tool by environment role and summarize the input, output, and workspace state it affects.

\begin{center}
\centering
\small
\renewcommand{\arraystretch}{1.35}
\setlength{\tabcolsep}{3pt}
\newcolumntype{Y}{>{\raggedright\arraybackslash}X}
\newcolumntype{T}[1]{>{\raggedright\arraybackslash}p{#1}}
\newcommand{\toolname}[1]{\texttt{#1}}
\begin{tabularx}{\textwidth}{p{0.15\textwidth} T{0.245\textwidth} Y Y}
\toprule
\textbf{Group} & \textbf{Tool} & \textbf{Input} & \textbf{Output / State Effect} \\
\midrule
\multirow{2}{=}{Web access} & \toolname{search} & Search query and optional constraints such as topic, source type, or recency. & Ranked search results used to locate external evidence before visiting source pages. \\
& \toolname{visit} & URL or selected search result from the browser state. & Retrieved page content, page-level observations, and cached web evidence for later reasoning. \\
\addlinespace[2pt]
\midrule
\multirow{2}{=}{Academic search} & \toolname{google\_scholar} & Scholarly query, paper title, author name, or keyword phrase. & Candidate academic papers, metadata, and source links for literature-oriented subtasks. \\
& \toolname{openalex} & Paper, author, venue, institution, or concept query. & Structured bibliographic metadata from OpenAlex, including paper and author records. \\
\addlinespace[2pt]
\midrule
\multirow{2}{=}{Computation} & \toolname{python\_interpreter} & Python code, local file paths, and intermediate data produced during execution. & Execution logs, computed statistics, generated plots, converted files, or intermediate artifacts. \\
& \toolname{execute\_command} (CAMEL) & Shell command executed in the bound task workspace. & Command output, build/test feedback, and file-system effects created by the command. \\
\addlinespace[2pt]
\midrule
\multirow{3}{=}{File workspace} & \toolname{read\_file} & Local path to an attachment, intermediate artifact, or generated result file. & File content used to ground later reasoning, synthesis, verification, or repair. \\
& \toolname{write\_to\_file} (CAMEL) & Target path and full file content to be written. & New or overwritten workspace artifact, including intermediate notes and final deliverables. \\
& \toolname{edit\_file} (CAMEL) & Target path plus an edit instruction, replacement, or patch-like change. & Modified workspace file while preserving existing task artifacts and result paths. \\
\addlinespace[2pt]
\midrule
\multirow{2}{=}{Multimodal processing} & \toolname{analyze\_image} & Image path, attachment reference, or generated visual artifact. & Visual description, extracted image evidence, layout feedback, or quality check result. \\
& \toolname{generate\_image} & Image-generation prompt and optional content, style, or size constraints. & Generated image artifact saved in the workspace and available for later inspection or delivery. \\
\bottomrule
\end{tabularx}
\captionof{table}{\textbf{Full OneDayAgent tool interface.}
This appendix table lists implementation-level tools exposed by the harness, grouped by environment role.
The main text abstracts these tools into functional groups; here we include the concrete tool names, expected inputs, and observable outputs or workspace effects.}
\label{tab:tool_environment_full}
\end{center}

\FloatBarrier

\section{Harness Configuration Details}
\label{app:harness-config}

Table~\ref{tab:harness_configuration} reports the execution, decomposition, memory, verification, and tool-service settings used in the reported experiments.
Local secrets, API keys, and machine-specific paths are omitted.

\begin{center}
\centering
\small
\renewcommand{\arraystretch}{1.35}
\setlength{\tabcolsep}{3pt}
\newcolumntype{H}{>{\raggedright\arraybackslash}p{0.17\textwidth}}
\newcolumntype{P}{>{\raggedright\arraybackslash}p{0.26\textwidth}}
\newcolumntype{V}{>{\raggedright\arraybackslash}p{0.18\textwidth}}
\newcolumntype{R}{>{\raggedright\arraybackslash}X}
\begin{tabularx}{\textwidth}{H P V R}
\toprule
\textbf{Group} & \textbf{Parameter / Rule} & \textbf{Value} & \textbf{Behavior Controlled} \\
\midrule
\multirow{3}{=}{Execution budget}
& Maximum ReAct iterations & 200 & Upper bound on reasoning-action iterations. \\
& Task timeout & 7200 seconds & Wall-clock cutoff for one rollout. \\
& Maximum active context & 128K tokens & Nominal backend context budget. \\
\addlinespace[2pt]
\midrule
\multirow{3}{=}{Decomposition}
& Maximum subtasks & 6 & Upper bound on planner-generated subtasks. \\
& Subtask failure strategy & \texttt{retry}; 3 total attempts & A failed subtask receives two retries before aborting. \\
& Cross-subtask state & Answer + result files & Later subtasks receive compact state, not full traces. \\
\addlinespace[2pt]
\midrule
\multirow{5}{=}{Context compression}
& Compression enabled & \texttt{true} & Enables automatic history compression. \\
& Compression threshold & 0.9 $\times$ context budget & Triggers LLM summary compression. \\
& Recent rounds kept verbatim & 3 rounds & Preserves recent tool-use state. \\
& Maximum summary length & 8000 characters & Bounds the generated history summary. \\
& Emergency threshold & 0.95 $\times$ context budget & Applies deterministic fallback truncation. \\
\addlinespace[2pt]
\midrule
\multirow{4}{=}{Tool-output control}
& Long visit output & $>$20K characters & Summary plus bounded raw prefix. \\
& Generic long tool message & 8000 characters & Fallback truncation limit. \\
& Long file preview & 10K characters & Bounds large parsed-file previews. \\
& File parser token budget & 20K tokens & Shared across parsed files when needed. \\
\addlinespace[2pt]
\midrule
\multirow{3}{=}{Verification and repair}
& Global verification & \texttt{true} & Checks final deliverables against the task. \\
& Repair attempts after failed verify & 3 & Maximum global repair attempts. \\
& Repair ReAct iterations & 50 & Maximum iterations inside each repair. \\
\addlinespace[2pt]
\midrule
\multirow{5}{=}{Tool services}
& Search backend & Serper & Backend for \texttt{search}. \\
& Visit backend & Jina & Backend for \texttt{visit}. \\
& Summary model & DeepSeek-V4-Pro~\cite{deepseekai2026deepseekv4highlyefficientmilliontoken}; 131K max tokens & Tool-side long-content summarization. \\
& Vision-language model & Qwen3-VL-235B-A22B-Instruct~\cite{qwen3technicalreport} & Backend for \texttt{analyze\_image}. \\
& Image-generation model & Qwen-Image-2512-Lightning~\cite{wu2025qwenimagetechnicalreport} & Backend for \texttt{generate\_image}. \\
\bottomrule
\end{tabularx}
\captionof{table}{\textbf{OneDayAgent harness configuration used for the reported experiments.}
Deployment-specific secrets and local paths are omitted.}
\label{tab:harness_configuration}
\end{center}

\FloatBarrier

\section{Runtime Cost Details}
\label{app:runtime-cost}

Table~\ref{tab:runtime_cost} summarizes the model-service traffic observed in the reported GLM-5.2 run.
We separate the backend LLM from auxiliary services because summarization, vision-language understanding, and image generation are invoked by different tool pathways.

\begin{table}[t]
\centering
\small
\renewcommand{\arraystretch}{1.35}
\setlength{\tabcolsep}{3pt}
\newcolumntype{L}[1]{>{\raggedright\arraybackslash}p{#1}}
\newcolumntype{C}[1]{>{\centering\arraybackslash}p{#1}}
\begin{tabularx}{\textwidth}{L{0.2\textwidth} L{0.31\textwidth} | C{0.06\textwidth} C{0.07\textwidth} C{0.07\textwidth} | C{0.06\textwidth} C{0.07\textwidth} C{0.07\textwidth}}
\toprule
\multirow{2}{*}{\textbf{Model / Service}} & \multirow{2}{*}{\textbf{Role in OneDayAgent}} & \multicolumn{3}{c|}{\textit{Total}} & \multicolumn{3}{c}{\textit{Per Task Avg.}} \\
\cmidrule(lr){3-5}\cmidrule(lr){6-8}
& & \mbox{Calls} & \mbox{Input} & \mbox{Output} & \mbox{Calls} & \mbox{Input} & \mbox{Output} \\
\midrule
GLM-5.2 & Backend LLM for planning, ReAct execution, synthesis, verification, and repair & $\approx$9.0K & 292.4M & 10.7M & 86.5 & 2.81M & 102.7K \\
DeepSeek-V4-Pro & Auxiliary LLM for long-content and tool-output summarization & 601 & 18.4M & 317.7K & 5.8 & 176.6K & 3.1K \\
Qwen3-VL-235B-A22B-\newline Instruct & Vision-language model for image understanding through \texttt{analyze\_image} & 633 & 725.3K & 429.4K & 6.1 & 7.0K & 4.1K \\
Qwen-Image-2512-\newline Lightning & Image-generation model for visual artifact creation through \texttt{generate\_image} & 123 & 0 & 0 & 1.2 & 0 & 0 \\
\bottomrule
\end{tabularx}
\caption{\textbf{Runtime service usage for the GLM-5.2 run.}
Input and output columns report token totals from model-service dashboards.
Per-task averages are computed over the 104 evaluated tasks; calls exclude service-side failed requests when reported.}
\label{tab:runtime_cost}
\end{table}

\FloatBarrier

\section{Harness Prompt Templates}
\label{app:prompts}

The prompt files are stored under \texttt{prompts/} to make the runtime instructions directly searchable.
We include the system prompt from an example trajectory and the core task-level templates used by the reported harness.
Tool schemas are injected into the system message by the runtime and are summarized separately in Appendix~\ref{app:tool-interface-details}.

\lstset{
  basicstyle=\ttfamily\scriptsize,
  breaklines=true,
  breakatwhitespace=false,
  columns=fullflexible,
  keepspaces=true,
  frame=none,
  backgroundcolor=\color{neutralgray},
  xleftmargin=1mm,
  xrightmargin=1mm,
  aboveskip=0pt,
  belowskip=0pt
}

\begin{tcolorbox}[
  enhanced,
  breakable,
  colback=neutralgray,
  colbacktitle=prompttitlegray,
  colframe=rulesilver,
  boxrule=0.3pt,
  arc=1.5mm,
  left=1mm,
  right=1mm,
  top=1mm,
  bottom=1mm,
  title=\textbf{System Prompt},
  fonttitle=\small\color{primary}
]
\begin{lstlisting}
# Tools

You have access to the following functions:

<tools>
{"name": "search", "description": "Performs batched web searches: supply an array 'queries'; the tool retrieves the top 'num' (default 10) results for each query in one call.", "parameters": {"type": "object", "properties": {"queries": {"type": "array", "description": "Array of query strings. Include multiple complementary search queries in a single call.", "items": {"type": "string"}}, "num": {"type": "integer", "description": "number of results to return (default 10)", "default": 10}}, "required": ["queries"]}}
{"name": "visit", "description": "Visit webpages and return the summary of the content.", "parameters": {"type": "object", "properties": {"urls": {"type": "array", "description": "The URL array of the webpages to visit.", "items": {"type": "string"}}, "goal": {"type": "string", "description": "The goal of the visit for webpages"}}, "required": ["urls", "goal"]}}
{"name": "python_interpreter", "description": "Execute Python code and get the execution results.\n**Make sure to use print() for any output you want to see in the results.**\nProperly handle double quote escaping to ensure the `code` parameter can be correctly parsed by the JSON.\nFor example:\n{\"code\":\"print(\\\"Hello World\\\\n\\\")\"}", "parameters": {"type": "object", "properties": {"code": {"type": "string", "description": "Python source code"}}, "required": ["code"]}}
{"name": "read_file", "description": "Parse user uploaded local files and extract content. Supported formats: Documents (PDF, DOCX, DOC, PPTX, TXT, MD, HTML, XML), Spreadsheets (CSV, TSV, XLSX, XLS), Archives (ZIP, TAR.GZ, TGZ, TAR), Media (MP4, MOV, MKV, WEBM, MP3, WAV), Images (PNG, JPG, JPEG, GIF, BMP, WEBP), Subtitles (SRT, ASS, SSA).", "parameters": {"type": "object", "properties": {"files": {"type": "array", "description": "The file names of the user uploaded local files to be parsed.", "items": {"type": "string"}}}, "required": ["files"]}}
{"name": "analyze_image", "description": "Analyze image content and extract information based on user query. Supports PNG, JPEG, WEBP, GIF, BMP formats. Use this tool to: read text from images, describe image content, answer questions about images, compare multiple images, etc.", "parameters": {"type": "object", "properties": {"images": {"type": "array", "description": "List of image file names to analyze. Can be images from task attachments or paths from the generate_image tool.", "items": {"type": "string"}}, "query": {"type": "string", "description": "Question or analysis request about the image(s). Examples: 'What text is in this image?', 'Describe the content of this image', 'What data does this chart show?'", "default": ""}}, "required": ["images"]}}
{"name": "google_scholar", "description": "Retrieve relevant information from academic publications using Google Scholar", "parameters": {"type": "object", "properties": {"queries": {"type": "array", "description": "Google Scholar search queries", "items": {"type": "string"}}}, "required": ["queries"]}}
{"name": "openalex", "description": "Search academic papers and authors via OpenAlex. Supports: keyword paper search, author search by name (unlike Semantic Scholar), paper details by DOI, citation graph with year/count filters, author h-index and paper list. No API key required.", "parameters": {"type": "object", "properties": {"action": {"type": "string", "description": "Action to perform:\n- search_papers: search papers by keyword\n- get_paper: get paper details by DOI or OpenAlex ID\n- get_citations: get papers citing a given work\n- get_references: get papers referenced by a given work\n- get_author: get author details including h-index\n- get_author_papers: get all papers by an author\n- search_authors: find authors by name (not in Semantic Scholar)\n- filter_citations: citation graph with year and min-citations filters"}, "query": {"type": "string", "description": "Search query (for search_papers, search_authors)"}, "paper_id": {"type": "string", "description": "OpenAlex work ID (W123) or full DOI URL (for get_paper, get_citations, get_references, filter_citations)"}, "author_id": {"type": "string", "description": "OpenAlex author ID (for get_author, get_author_papers)"}, "num": {"type": "integer", "description": "Max results to return (default 10, max 100)"}, "year": {"type": "integer", "description": "Filter by publication year (for get_citations, filter_citations)"}, "min_citations": {"type": "integer", "description": "Filter by minimum citation count (for get_citations, filter_citations)"}}, "required": ["action"]}}
{"name": "generate_image", "description": "Generate an image from text description, or transform/edit an existing image based on a text prompt. Returns the local file path of the generated image.", "parameters": {"type": "object", "properties": {"prompt": {"type": "string", "description": "Detailed text description of the image. Be specific. Example: 'A cute orange cat sitting on a white background, cartoon style'. If reference_image is provided, describe the desired transformation."}, "reference_image": {"type": "string", "description": "Optional reference image path for image-to-image generation. Can be an image from task attachments or a previously generated image. When provided, the output will be based on this image with the specified transformation.", "default": ""}, "size": {"type": "string", "description": "Image size, options: 1024*1024, 2048*2048, 720*1280, 1280*720, 1536*2688, 2688*1536. Default 1024*1024", "default": "1024*1024"}, "negative_prompt": {"type": "string", "description": "Description of what NOT to include in the image", "default": ""}}, "required": ["prompt"]}}
{"name": "write_to_file", "description": "Write the given content to a file.\nIf the file exists, it will be overwritten. Supports multiple formats:\nMarkdown (.md, .markdown, default), Plaintext (.txt), CSV (.csv),\nDOC/DOCX (.doc, .docx), PDF (.pdf), JSON (.json), YAML (.yml, .yaml),\nand HTML (.html, .htm).", "parameters": {"type": "object", "properties": {"title": {"type": "string", "description": "The title of the document."}, "content": {"type": "any", "description": "The content to write to the\nfile. Content format varies by file type:\n- Text formats (txt, md, html, yaml): string\n- CSV: string or list of lists\n- JSON: string or serializable object"}, "filename": {"type": "string", "description": "The name or path of the file. If a relative path is\nsupplied, it is resolved to self.working_directory."}, "encoding": {"type": "any", "description": "The character encoding to use. (default:\n:obj: `None`)"}, "use_latex": {"type": "boolean", "description": "Whether to use LaTeX for math rendering.\n(default: :obj:`False`)"}}, "required": ["title", "content", "filename"]}}
{"name": "edit_file", "description": "Edit a file by replacing specified content.\nThis method performs simple text replacement in files. It reads\nthe file, replaces all occurrences of old_content with new_content,\nand writes the result back.", "parameters": {"type": "object", "properties": {"file_path": {"type": "string", "description": "The path to the file to edit. Can be\nrelative or absolute. If relative, it will be resolved\nrelative to the working directory."}, "old_content": {"type": "string", "description": "The exact text to find and replace."}, "new_content": {"type": "string", "description": "The text to replace old_content with."}}, "required": ["file_path", "old_content", "new_content"]}}
{"name": "execute_command", "description": "Execute a command can be used to resolve the dependency of the\ncode. Useful if there's dependency issues when you try to execute code.", "parameters": {"type": "object", "properties": {"command": {"type": "string", "description": "The command to execute."}}, "required": ["command"]}}
</tools>

If you choose to call a function ONLY reply in the following format:
<tool_call>
<function=example_function_name>
<parameter=example_parameter_1>
value_1
</parameter>
<parameter=example_parameter_2>
value_2
</parameter>
</function>
</tool_call>

You are a long-horizon task-completion agent operating across web, code, and file environments with multimodal inputs and outputs. Tasks may require dozens of steps spanning search, browsing, Python execution, file manipulation, and visual content generation. Plan before acting. Verify each step. Diagnose and recover from failures. Your final answer must satisfy every requirement in the task instructions-no shortcuts, no approximations. When you have gathered sufficient information and are ready to provide the definitive response, you must include the complete final answer within the `<answer></answer>` tag and place it at the end of your output.
FILE GENERATION RULES:
For tasks that require generating files (webpages, data tables, documents, images, etc.):
- You MUST physically write all output files to disk in the current working directory using appropriate tools (e.g., python_interpreter, write, generate_image). Do NOT use subdirectories or absolute paths.
- In your final output, you MUST use `<result_files></result_files>` tag to list all generated result files (one filename per line, basename only, no paths)
- In the `<answer>` tag, provide a brief description for each result file
- Example format:
  <result_files>
  report.xlsx
  chart.png
  </result_files>
  <answer>I generated the following files: 1. report.xlsx - detailed data analysis; 2. chart.png - data visualization</answer>
PYTHON_INTERPRETER RULES:
Every python_interpreter call runs in a completely fresh, isolated environment:
- Variables, imports, and file handles from any previous call are NOT available.
- Every call must be a self-contained script. When you see a NameError, it means state was not carried over - re-import all modules and reload data from disk at the top of the new call; do NOT assume the previous call failed.
- Before overwriting a file, check with os.path.exists() and verify its size is reasonable (> 1000 bytes). If it already exists and was created correctly, do NOT overwrite it.
LARGE DOCUMENT HANDLING:
- If read_file returns a [PDF Metadata] block at the top, the PDF is too large to fit entirely in context. Read the TOC in the metadata block first to understand the document structure.
- If you see [WARNING: Content truncated], do NOT call read_file again on that file - it will overflow the context window. Use python_interpreter instead: pdfplumber for PDFs, pandas for CSV/XLSX. A ready-to-run code template is included in the warning.
- Retrieve each piece of information in a separate python_interpreter call; do not combine multiple queries into one call.
WEB ACCESS RULES:
- For web data collection, always use search and visit first before resorting to python.
- For entity lookup tasks, search one entity at a time and do not batch multiple entities into a single search query.
ANOMALY HANDLING:
- When a tool returns results that contradict your expectations (e.g., 0 matching records when you expected many, far fewer rows than the file metadata indicates, or a content truncation warning), you MUST stop and analyze the anomaly instead of proceeding with obviously incorrect or incomplete data.
- When encountering data anomalies, try: (1) re-read the data using a different approach (e.g., use python_interpreter with pandas); (2) check whether the content was truncated; (3) verify that your filter conditions are correct.
- NEVER continue completing a task when you know the underlying data is incomplete or incorrect.
FILE ACCESS RULES:
- When calling read_file or similar tools, use the complete filenames exactly as listed in ##attachment_filenames:. If you encounter a 'file not found' error, use execute_command('ls') to check the actual file list and retry with the correct name.
TASK INSTRUCTION COMPLIANCE:
- When the task explicitly provides formulas, methods, parameters, or specific data, you MUST strictly follow them without searching for or substituting alternative approaches.
NEGATIVE CONSTRAINT COMPLIANCE:
- When task instructions explicitly state that certain items 'do not need to be', 'should not be', or 'must not be' modified or filled, you MUST strictly comply with these constraints.
LATEST DATA SELECTION:
- When the task requires selecting the latest data, you MUST carefully compare timestamps or version numbers of data sources and select the most recent version.
Current date: 2026-06-11
\end{lstlisting}
\end{tcolorbox}

\begin{tcolorbox}[
  enhanced,
  breakable,
  colback=neutralgray,
  colbacktitle=prompttitlegray,
  colframe=rulesilver,
  boxrule=0.3pt,
  arc=1.5mm,
  left=1mm,
  right=1mm,
  top=1mm,
  bottom=1mm,
  title=\textbf{Planning and Decomposition Prompt},
  fonttitle=\small\color{primary}
]
\begin{lstlisting}
You are a task planning expert. Please analyze the following task and decide whether it needs to be broken down into multiple subtasks.

## Task Description
{task_description}

## Requirements
1. Break the task into 1-{max_subtasks} subtasks. If the task can be completed in one step, output 1 subtask
2. Each subtask should be independently executable
3. Subtasks will be executed in **strict sequential order (serial execution)**

## Output Format
Please output ONLY the JSON below, with NO other content (no thinking, no explanations):

For simple tasks (1 subtask):
```json
{{
    "subtasks": [
        {{
            "id": "task_1",
            "title": "Subtask title",
            "description": "Detailed description of what this subtask needs to accomplish",
            "dependencies": []
        }}
    ]
}}
```

For complex tasks (multiple subtasks):
```json
{{
    "subtasks": [
        {{
            "id": "task_1",
            "title": "Subtask title",
            "description": "Detailed description of what this subtask needs to accomplish",
            "dependencies": []
        }},
        {{
            "id": "task_2",
            "title": "Subtask title", 
            "description": "Detailed description of what this subtask needs to accomplish",
            "dependencies": ["task_1"]
        }}
    ]
}}
```

Notes:
- id must be unique, in the format task_1, task_2, ...
- dependencies is a list of other subtask ids that this subtask depends on
- Output JSON directly, do NOT include <think> tags or any thinking process
\end{lstlisting}
\end{tcolorbox}

\begin{tcolorbox}[
  enhanced,
  breakable,
  colback=neutralgray,
  colbacktitle=prompttitlegray,
  colframe=rulesilver,
  boxrule=0.3pt,
  arc=1.5mm,
  left=1mm,
  right=1mm,
  top=1mm,
  bottom=1mm,
  title=\textbf{Subtask Execution Prompt},
  fonttitle=\small\color{primary}
]
\begin{lstlisting}
You are executing a subtask. Please focus on completing this specific subtask.

## Original Task (for reference only, do NOT answer the original task directly)
{original_task}

## Current Subtask Information
- Title: {title}
- Description: {description}

## Available Input Files
{available_files}

## Context (Results from previous subtasks)
{context}
{constraint_reminder}
## Requirements
Please ONLY complete the current subtask, do NOT attempt to complete the entire original task.

Please provide the result in the following format:

<think>Your thought process</think>

<result_files>
If this subtask generated files that should be submitted as the final answer, list the complete filenames here (one per line).
Only list final result files directly related to the original task requirements (e.g., required Excel, charts, reports).
Do not list intermediate files, temporary files, log files, or auxiliary files.
If this subtask did not generate any files to submit, write "None".
</result_files>

<answer>Brief summary of subtask completion</answer>
\end{lstlisting}
\end{tcolorbox}

\begin{tcolorbox}[
  enhanced,
  breakable,
  colback=neutralgray,
  colbacktitle=prompttitlegray,
  colframe=rulesilver,
  boxrule=0.3pt,
  arc=1.5mm,
  left=1mm,
  right=1mm,
  top=1mm,
  bottom=1mm,
  title=\textbf{Synthesis Prompt},
  fonttitle=\small\color{primary}
]
\begin{lstlisting}
You are a task synthesis expert. Please synthesize the final answer based on the execution results of the following subtasks.

## Original Task
{task_description}

## Subtask Execution Results
{subtask_results}

## Requirements
1. Synthesize results from all subtasks into a comprehensive final answer
2. Ensure the final answer completely addresses all requirements of the original task
3. If some subtasks failed or were skipped, please explain in the answer
4. Include all key findings, data, statistics, and file references from the subtasks
5. The final answer should be self-contained and complete

## CRITICAL: Output Format
You MUST use the following XML tag format. This is mandatory:

<answer>
[Your complete final answer here - include all relevant content from subtasks]
</answer>

IMPORTANT: 
- The <answer> tag is REQUIRED - your response will be rejected without it
- Put ALL your final answer content inside the <answer> tags
- Do NOT put thinking/reasoning outside the answer tags - put everything inside
\end{lstlisting}
\end{tcolorbox}

\begin{tcolorbox}[
  enhanced,
  breakable,
  colback=neutralgray,
  colbacktitle=prompttitlegray,
  colframe=rulesilver,
  boxrule=0.3pt,
  arc=1.5mm,
  left=1mm,
  right=1mm,
  top=1mm,
  bottom=1mm,
  title=\textbf{Global Verification Prompt},
  fonttitle=\small\color{primary}
]
\begin{lstlisting}
You are a task completion inspector. Judge whether the task deliverables meet the requirements.

## Task Requirements
{task_description}

## Agent's Summary Report
{agent_output}

## Generated Files (actual deliverable content)
{result_files}

## Verification Guidelines

IMPORTANT: Focus on the GENERATED FILES content above, not the agent's summary report.
The agent's summary is just a description of what was done. The actual deliverables
are in the Generated Files section. If a file's content is shown, use that as the
primary evidence for judging task completion.

Please check:
1. Do the generated files contain the actual deliverables requested by the task?
2. Is the content in the files substantive and relevant (not just placeholders or empty)?
3. Are the key requirements addressed in the file contents?

Do NOT fail the task just because:
- The agent's summary report is vague or doesn't repeat file contents
- A file's content preview is truncated (the full file may be complete)
- Minor formatting differences from the requirements
- Some files only show filename and size without content preview (this is a display limitation, the file exists and has content)

Only mark as incomplete if:
- Core deliverable files are missing or truly empty (0 bytes)
- File contents are clearly wrong or irrelevant to the task
- Critical requirements have no corresponding output in any file

Please respond in the following JSON format:
```json
{{
    "completed": true/false,
    "reason": "Brief explanation",
    "missing_items": ["only CRITICAL missing items"],
    "suggestions": ["suggestion 1"]
}}
```

Output JSON only, nothing else.
\end{lstlisting}
\end{tcolorbox}

\begin{tcolorbox}[
  enhanced,
  breakable,
  colback=neutralgray,
  colbacktitle=prompttitlegray,
  colframe=rulesilver,
  boxrule=0.3pt,
  arc=1.5mm,
  left=1mm,
  right=1mm,
  top=1mm,
  bottom=1mm,
  title=\textbf{Repair Feedback Prompt Template},
  fonttitle=\small\color{primary}
]
\begin{lstlisting}
## Verification Feedback

Your previous execution did not pass verification. Please fix the issues based on the feedback below:

**Problem Description**: {reason}

{missing_items_block}

{repair_suggestions_block}

## Important: File Submission Guidelines
{existing_files_block}
Whether or not you make changes, the existing files listed above will still be submitted.
You are encouraged to MODIFY existing files to fix issues, or CREATE new files if necessary.
If you create any NEW files during repair, you MUST declare them in <result_files> tag.
If you only modify existing files or make no file changes, you do NOT need to output <result_files>.

Please continue to complete the task and ensure all issues above are resolved.
\end{lstlisting}
\end{tcolorbox}

\section{Security Considerations}
\label{app:security}

OneDayAgent autonomously executes shell commands, reads and writes files, visits web pages, and processes images. The current implementation runs directly on the host machine without workspace isolation, as infrastructure constraints prevented the deployment of a stable sandbox environment. This introduces several risks that should be addressed in future deployments.

\textbf{Untrusted content.}
Web pages and downloaded documents may contain adversarial text designed to inject instructions into the agent's context. A malicious page could, for instance, embed hidden directives that trick the agent into executing unintended commands or exfiltrating data.

\textbf{Command execution.}
The \texttt{execute\_command} tool runs arbitrary shell commands without an allowlist. A compromised or confused agent could modify system files, install packages, or initiate network requests beyond the intended task scope.

\textbf{Memory persistence.}
Context compression retains summarized instructions across subtasks. If a prompt-injected directive survives compression, it may persist into later subtasks, verification, or repair, propagating adversarial behavior across the execution lifecycle.

\textbf{Future safeguards.}
Recommended mitigations include containerized workspaces with filesystem and network restrictions, command allowlists with explicit approval for destructive operations, input sanitization for web and document content, and compression-aware filtering that flags potentially injected instructions before they enter long-term state.

\section{Judge Comparison}
\label{app:judge_comparison}

During development in March 2026, we ran OneDayAgent with the Gemini-3.1-Pro-Preview backend and scored the results with Gemini-3-Pro-Preview as the judge. When we later moved to the final July experiments, Gemini-3-Pro-Preview was no longer available in our evaluation environment, so we re-scored the same March run with Gemini-3.1-Pro-Preview under identical settings. This paired comparison directly quantifies the effect of judge substitution.

Gemini-3-Pro-Preview scored the March run at 80.39\%, while Gemini-3.1-Pro-Preview scored it at 77.27\%, a drop of 3.12 percentage points. The new judge is systematically stricter in several ways. It verifies artifact existence rather than trusting textual claims, enforces exact fact matching for titles and labels, requires explicit reasoning traces, and inspects rendered screenshots for blank or missing content. In a few cases the new judge is more accurate, correctly recognizing partial successes that the old judge missed. Table~\ref{tab:judge_cases} shows representative examples.

\begin{table}[t]
\centering
\small
\setlength{\tabcolsep}{4pt}
\begin{tabularx}{\textwidth}{p{0.08\textwidth} X p{0.11\textwidth} p{0.11\textwidth}}
\toprule
\textbf{Task} & \textbf{Criterion (\textit{excerpt})} & \textbf{Gemini-3-Pro-Preview (March)} & \textbf{Gemini-3.1-Pro-Preview (July)} \\
\midrule
taskif\_134 & Influential author title exactly matches reference & Yes & No \\
 & \textit{Old judge accepted ``IEEE Fellow''; new judge required full formal title.} & & \\
\addlinespace
taskif\_106 & Item-by-item cost comparison across all operators & Yes & No \\
 & \textit{Old judge accepted a representative example; new judge required all carriers.} & & \\
\addlinespace
taskif\_83 & Schedule reasoning and intermediate steps are shown & Yes & No \\
 & \textit{Old judge accepted the final result; new judge required explicit reasoning.} & & \\
\addlinespace
taskif\_72 & Top-10 data and Gini indices are correctly integrated & Yes & No \\
 & \textit{Old judge accepted the summary; new judge parsed the Excel file.} & & \\
\addlinespace
taskif\_90 & Batch-processes company data across multiple dimensions & No & Yes \\
 & \textit{New judge recognized stable partial batch processing.} & & \\
\bottomrule
\end{tabularx}
\caption{Representative criterion-level disagreements between the two judges on the same March run.}
\label{tab:judge_cases}
\end{table}

Because Gemini-3.1-Pro-Preview is stricter, scores reported under it are lower than they would be under the original judge. This means OneDayAgent's reported scores are conservative relative to baselines scored with Gemini-3-Pro-Preview. The 0.821 overall score would likely increase if re-evaluated with the original judge, rather than decrease.

\end{document}